\documentclass[letterpaper]{article} 
\usepackage[preprint]{aaai2027}  
\usepackage[hyphens]{url}  
\usepackage{graphicx} 
\usepackage{natbib}  
\usepackage{caption} 
\usepackage{algorithm}
\usepackage{algorithmic}

\usepackage{newfloat}
\usepackage{listings}
\DeclareCaptionStyle{ruled}{labelfont=normalfont,labelsep=colon,strut=off} 
\floatstyle{ruled}
\newfloat{listing}{tb}{lst}{}
\floatname{listing}{Listing}

\usepackage{booktabs}

\usepackage{CJKutf8}
\usepackage{graphicx} 
\usepackage{booktabs}
\usepackage{multibib}
\usepackage{caption}
\usepackage{subcaption}
\usepackage{multirow}
\usepackage[accsupp]{axessibility}
\usepackage{amsthm,amsmath,amssymb}
\usepackage{pifont}
\usepackage{mathrsfs}
\usepackage{bbding}
\usepackage{url}            %
\usepackage[table]{xcolor}

\newcommand{\app}{\raise.17ex\hbox{$\scriptstyle\sim$}}

\definecolor{baselinecolor}{gray}{.9} 
\newcommand{\baseline}[1]{\cellcolor{baselinecolor}#1}

\newcolumntype{*}{>{\global\let\currentrowstyle\relax}}
\newcolumntype{^}{>{\currentrowstyle}}

\definecolor{dt}{gray}{0.7}  %

\newcommand{\ie}{{\emph{i.e.}}, }

\newcommand{\eg}{{\emph{e.g.}}, }
\newcommand{\etc}{etc.}

\usepackage[capitalize]{cleveref}
\crefname{section}{Sec.}{Secs.}
\Crefname{section}{Section}{Sections}
\Crefname{table}{Table}{Tables}
\crefname{table}{Tab.}{Tabs.}

\newcolumntype{S}{@{}>{\lrbox0}l<{\endlrbox}}  %
\definecolor{lightgreen}{HTML}{D8ECD1}

\definecolor{deemph}{gray}{0.6}

\usepackage{balance}

\title{Sign Language Question Answering: \\A New Task, Benchmark, and Baseline for Sign Language Understanding}

\author{
    Shiwei Gan\equalcontrib, Lichen Wang\equalcontrib, Xiao Liu, Yafeng Yin\corresponding, Kuizhuang Liu, Sanglu Lu, Lei Xie 
}
\affiliations{
    State Key Laboratory of Novel Software Technology, Nanjing University, Nanjing 210023, China \\
    sw@nju.edu.cn, \{wanglichen, liuxiaox\}@smail.nju.edu.cn, yafeng@nju.edu.cn \\
    liukz@smail.nju.edu.cn, \{sanglu,lxie\}@nju.edu.cn
}

\begin{document}
\maketitle

Recent advances in sign language (SL) understanding (SLU) have led to remarkable progress in tasks such as continuous SL recognition and SL translation. However, these tasks are designed with predefined objectives, requiring models to learn a fixed mapping from sign videos to glosses or spoken-language sentences. As a result, they provide only a limited assessment of whether a model truly understands the semantic content of SL videos. 
To address this limitation, \textbf{we first propose a new task, Sign Language Question Answering (SLQA)}, which evaluates SL understanding by requiring models to answer natural language questions about SL videos. Unlike previous SLU tasks, SLQA provides a more flexible and comprehensive evaluation framework that assesses multiple reasoning capabilities beyond recognition and translation.
To facilitate this task, \textbf{we further construct two SignQA benchmarks} based on PHOENIX14T and CSL-Daily by automatically generating question-answer pairs from existing gloss and sentence annotations using carefully designed templates. The resulting datasets cover five complementary question categories, including position reasoning, structural reasoning, visual search, gloss recognition, and translation understanding.
\textbf{Finally, we propose a simple yet effective baseline model} equipped with a Question-Conditioned Modulated Temporal Downsampling module and an in-domain knowledge transfer strategy, enabling effective knowledge transfer from existing SLU tasks while enhancing question-aware temporal feature modeling.
Extensive experiments demonstrate that our baseline consistently outperforms representative vision-language models across all question categories, establishing a strong benchmark for future research on SLQA. Datasets are available at:{https://huggingface.co/datasets/hulala/SignQA-2026}.

\section{Introduction}
Recent advances in Sign Language(SL) understanding(SLU) have led to significant progress across a wide range of research tasks, including isolated SL recognition (ISLR)~\cite{hu2021hand}, continuous SL recognition (CSLR)~\cite{min2021visual, gan2024signgraph}, gloss-based SL translation (GBSLT)~\cite{gan2025mixsigngraph, chen2022two}, gloss-free SL translation (GFSLT)~\cite{gan2026signllama}, \etc Specifically, ISLR focuses on classifying an isolated sign into a predefined sign category. Extending this setting, CSLR aims to recognize  a continuous SL video and convert it into a sequence of glosses. Building upon CSLR, GBSLT further translates SL videos into spoken sentences with the supervision of gloss annotations. To alleviate the reliance on costly gloss annotations, GFSLT seeks to directly train SLT models to translate SL videos into spoken text without using gloss supervision.

However, most of these tasks are task-specific and trained with fixed objectives, where models are optimized to produce predefined outputs (\eg glosses or text sequences). Although these approaches have advanced the mapping of SL videos to textual representations, they remain limited in terms of flexibility and generalization. In particular, (i) they struggle to accommodate diverse and open-ended user requirements, and (ii) their fixed objectives make it difficult to determine whether a model has genuinely learned to understand the semantic content of SL videos, rather than merely learning task-specific mappings or memorizing the training data.

\begin{figure}
    \centering
    \includegraphics[width=0.95\linewidth]{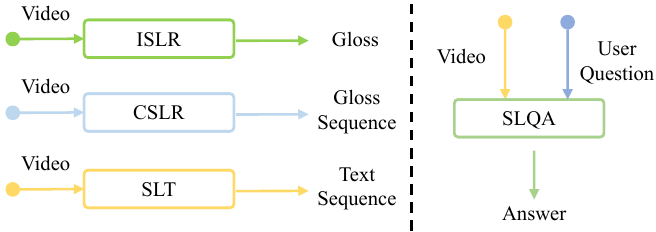}
  \caption{Illustration of existing SLU and proposed SLQA.}
   \vspace{-2mm}
    \label{fig:intor}
    \vspace{-4mm}
\end{figure}

To address these limitations, inspired by the success of Video Question Answering (VideoQA)~\cite{zhong2022video,wu2025videoqa, ataallah2024minigpt4,li2024llama}, (1) \textbf{we propose a new task termed SLQA}. Unlike existing SL tasks that focus on predefined objectives, such as sign classification, gloss recognition, or video-to-text translation, SLQA requires a model to answer diverse natural-language questions based on a SL video. Specifically, the model takes a SL video and a natural language question as input, and generates the corresponding answer. This formulation allows users to query different aspects of the video content, including spatial and temporal relations, gloss sequences, visual details, and sentence-level semantics, thereby providing a more comprehensive evaluation of SLU than conventional task-specific benchmarks. 
(2) Besides, to evaluate the proposed SLQA task, \textbf{we construct two SLQA datasets} based on widely used SL benchmarks, namely PHOENIX14T and CSL-Daily. Specifically, we leverage existing annotations, including gloss-level annotations and sentence-level text annotations, to automatically construct a large number of SLQA pairs via carefully designed question templates. This allows us to generate diverse and controllable QA samples without requiring additional manual annotation. 
(3) Furthermore, \textbf{we propose   SLQAM: a  baseline model for SLQA}. To enable effective learning with limited SLQA training data, we introduce two key strategies. First, to better enable the model to attend to question-relevant video regions, we design a  \textbf{Question-Conditioned Modulated Temporal Downsampling (QCMTD)} module, which explicitly aligns the question semantics with temporal sign video representations, allowing the model to focus on question-relevant spatio-temporal features for answer prediction. Second,  we propose domain knowledge transfer, where the model is pre-trained on related SL tasks, including CSLR and SLT, and subsequently fine-tuned on the SLQA datasets. We empirically observe that training the model from scratch on SLQA leads to suboptimal performance due to limited data scale, while pre-training significantly improves generalization. Our main contributions are summarized as follows:

\begin{itemize}
    \item We propose a new SLU task: \textbf{SL Question Answering (SLQA)}. Unlike existing SLU tasks with fixed objectives, SLQA focuses on understanding SL videos by answering natural language questions related to the video content.

    \item We further propose a baseline model for SLQA. To effectively model the relationship between SL videos and questions, we introduce two key components: (1) \textbf{QCMTD}, which replaces fixed temporal downsampling and dynamically modulates informative temporal segments conditioned on the input question, thereby enhancing question-relevant video representations; (2) \textbf{In-domain knowledge transfer training}, which leverages three stage training pipeline to transfer domain-specific knowledge into SLQA learning.

    \item We construct a benchmark for the SLQA task to evaluate the effectiveness of the proposed method. Extensive experiments across five sub-tasks demonstrate that our approach achieves strong and consistent performance improvements over existing VideoQA models.
\end{itemize}

\section{Related}

\paragraph{Sign Language Understanding.}
Existing research in SLU has primarily focused on a set of task-specific objectives, including isolated SL recognition (ISLR), continuous SL recognition (CSLR), and SL translation (SLT). ISLR classifies an isolated sign into a predefined category~\cite{zhou2025scaling, zuo2023natural}, while CSLR recognizes continuous SL videos as sequences of glosses. Building upon CSLR, SLT aims to translate SL videos into spoken-language sentences. GBSLT typically leverages gloss annotations as intermediate supervision, whereas GFSLT seeks to directly learn the alignment between  sign videos and textual semantics without gloss supervision~\cite{zhou2023gloss,gan2025mixsigngraph,guo2025bridging}.~\nocite{gueuwouetal2025shubert, gueuwou2025signmusketeers, gan2026learning}
Despite their advances, these tasks are fundamentally defined by fixed objectives: a model is trained to map an SL video to a predefined type of output. Such task-specific formulations provide limited flexibility for evaluating the diverse capabilities required to understand an SL video. In particular, they do not allow a model to flexibly answer different questions about the same video, making it difficult to assess whether the model can selectively retrieve and reason over different aspects of the visual content.~\nocite{min2025closer,gan2024signgraph, gan2025mixsigngraph, gan2026signllama, gan2026learning,rusttowards,tanzer2024youtube,liu2025scope,fish2025geo,zhao2024conditional,wang2025gloss,guo2025bridging,jiao2024visual,lin2023gloss,li2025uni,gueuwouetal2025shubert,wong2025signrep,jiang2024signclip,chen2024factorized,rust2024towards,uthus2023youtube,liang2024llava}

Recent studies have begun to incorporate questions into SLU. Representative works, such as SSL-SSAW~\cite{liu2026ssl} and QD-SLT~\cite{gao2024overcoming}, introduce questions as additional conditions for SLT. Although different questions may guide the model to focus on different semantic aspects of the same SL video, the ultimate objective remains translation generation. Thus, the question serves as an auxiliary condition for generation  rather than the target of video-grounded reasoning itself. SLU-2K~\cite{testa2026slu} further introduces question-answer pairs for SL understanding. However, its questions and answers are confined to the textual translation, and the task primarily evaluates comprehension of the translated sentence rather than understanding the visual content of the SL video.

In contrast, we formulate SL Question Answering (SLQA) as a bora SLU task in which questions are directly grounded in SL videos. Given an SL video and a natural-language question, the model must answer the question by retrieving and reasoning over the relevant visual content, rather than simply generating a fixed-form output or reasoning over a pre-existing translation. Our benchmark covers five complementary question categories, including \emph{Position Reasoning}, \emph{Structural Reasoning}, \emph{Visual Search}, \emph{Gloss Recognition}, and \emph{Translation Understanding}, which collectively evaluate different levels of SL understanding, from fine-grained visual retrieval and sequential reasoning to gloss recognition and sentence-level semantic understanding. Therefore, SLQA provides a more flexible evaluation paradigm for assessing whether a model can genuinely understand the content conveyed by an SL video.

\begin{figure*}
    \centering
    \includegraphics[width=0.95\linewidth]{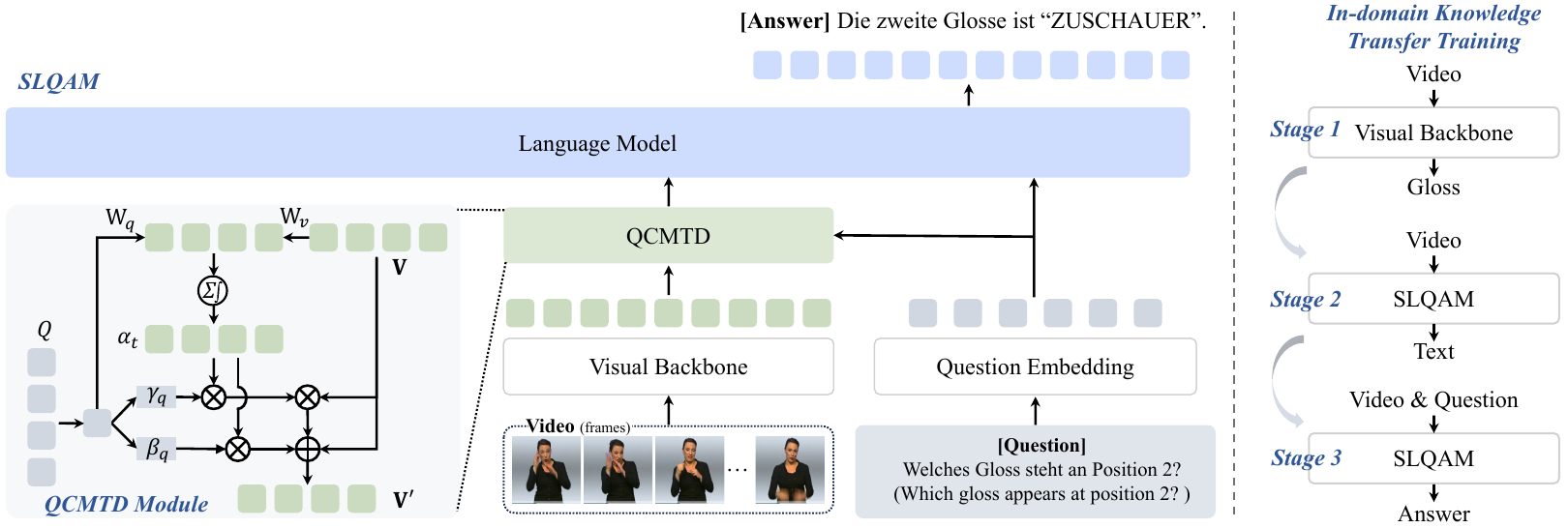}
        \vspace{-1mm}
    \caption{Overall architecture of the proposed SLQAM. The framework consists of a QCMTD module and an in-domain knowledge transfer module.}
    \label{fig:model}
    \vspace{-3mm}
\end{figure*}

\paragraph{Visual Question Answering.}
Visual Question Answering (VQA) lies at the intersection of computer vision and natural language processing, requiring models to understand visual content and perform reasoning over natural language queries. According to the modality of visual input, VQA can be broadly categorized into Image Question Answering (Image QA) and Video Question Answering (Video QA).

\noindent\textbf{Image Question Answering.}
Early Image QA methods primarily focused on learning joint visual–textual representations through attention mechanisms and multimodal fusion modules. Benchmark datasets such as VQA~\cite{antol2015vqa} and CLEVR~\cite{johnson2017clevr} have driven the development of bilinear pooling and co-attention networks for modeling fine-grained spatial–language interactions. With the emergence of Transformer architectures, Vision-Language Pretraining (VLP) models such as ViLBERT~\cite{lu2019vilbert}, BLIP~\cite{li2022blip}, and Flamingo~\cite{alayrac2022flamingo} further improved performance by leveraging large-scale image–text pairs to learn cross-modal alignment. More recently, Vision-Language Models (VLMs) and Multimodal Large Language Models (MLLMs), including LLaVA~\cite{liu2023visual} and GPT-4V, have shifted the paradigm toward instruction-tuned LLM-based reasoning, enabling open-ended and zero-shot visual question answering with strong generalization capability.

\noindent\textbf{Video Question Answering.}
Compared with Image QA, Video QA is inherently more challenging due to the additional temporal dimension, which requires reasoning over motion patterns, event ordering, and long-range temporal dependencies. Early approaches extended image-based architectures using RNNs, 3D CNNs(e.g., I3D), and temporal attention mechanisms to aggregate frame-level features, evaluated on benchmarks such as MSRVTT-QA~\cite{xu2017video} and ActivityNet-QA~\cite{yu2019activitynet}. To better model complex temporal structures, later methods introduced graph-based models and spatio-temporal Transformers, treating frames or object-level entities as nodes to capture cross-frame dependencies. In the current era of foundation models, adapting large language models to video understanding has become the dominant paradigm. Modern video-LLMs (\eg Video-LLaVA~\cite{lin2024video}, Video-ChatGPT~\cite{maaz2024video}, and LLaMA-VID~\cite{li2024llama}) typically rely on visual encoders combined with token compression or resampling modules to reduce the large number of video tokens while preserving essential temporal semantics.

Despite the remarkable progress of Video QA, existing methods are developed for natural videos and evaluated on open-domain benchmarks. SL videos, however, exhibit unique characteristics, including fine-grained hand articulations, non-manual markers, and strong temporal dependencies, which require substantially different visual representations and reasoning capabilities. Furthermore, unlike conventional Video QA that focuses on objects, events, and scene understanding, the proposed SLQA task requires reasoning over the linguistic content encoded in sign videos, including gloss recognition, structural reasoning, visual search, and translation understanding. Therefore, existing Video QA methods cannot be directly applied to this new setting.

\section{Method}

\paragraph{Overview.}
In this paper, we propose a new task, termed Sign Language Question Answering (SLQA). Specifically, given a SL video $V=\{f_t\}_{t=1}^{T}$ with $T$ frames and a natural language question $Q$, the objective of SLQA is to generate the corresponding answer $A$ conditioned on both the video and the question, i.e., $P(A\mid V,Q)$.
To facilitate research on this task, we first construct two SLQA datasets based on two widely used public SL datasets, PHOENIX14T and CSL-Daily. By leveraging the existing gloss and text annotations together with carefully designed question templates, we automatically generate diverse question-answer pairs covering five representative categories: \emph{Position Reasoning}, \emph{Structural Reasoning}, \emph{Visual Search}, \emph{Gloss Recognition}, and \emph{Translation Understanding}.
Furthermore, we develop a baseline model for SLQA. To effectively bridge SLU and question answering, we introduce two key designs: (1) a \emph{Question-Conditioned Modulated Temporal Downsampling} module that dynamically selects question-relevant video frames instead of using conventional fixed temporal downsampling, and (2) an \emph{in-domain knowledge transfer} strategy that transfers knowledge learned from CSLR and SLT to the SLQA task.  Finally, we establish a comprehensive benchmark on the proposed datasets and conduct extensive experiments to evaluate representative methods and the proposed baseline, providing a strong foundation for future research on SLQA.

\subsection{Dataset Construction.} 
\label{sec:dataset}
To facilitate the SLQA task, we construct two SLQA datasets based on two widely used SLT datasets, PHOENIX14T and CSL-Daily. Instead of collecting free-form question-answer pairs through manual annotation, we automatically generate QA pairs by leveraging the existing gloss annotations and spoken-language translations. Specifically, we design a collection of question templates that transform the original annotations into diverse question-answer pairs. This template-based construction strategy ensures that the generated answers are unambiguous, automatically verifiable, and scalable, while requiring no additional human annotation.

\paragraph{Question Categories.}

To comprehensively evaluate different aspects of SL understanding, we design five complementary question categories.

\begin{itemize}
\item \textbf{M1 -- Position Reasoning.}
Given a position in the gloss sequence (e.g., the first, last, middle, or $N$-th position), the model is required to identify the corresponding gloss.

\item \textbf{M2 -- Structural Reasoning.}
The task requires understanding the structural relationships within the gloss sequence, such as identifying neighboring glosses or the gloss located between two specified signs.

\item \textbf{M3 -- Visual Search.}
The model is required to localize specific visual content in a SL video, including retrieving the position of a target gloss or identifying non-manual markers (e.g., interrogative facial expressions). The latter is instantiated only for CSL-Daily since PHOENIX14T contains only declarative utterances.

\item \textbf{M4 -- Gloss Recognition.}
The model answers questions that require recognizing the gloss sequence corresponding to the input SL video.

\item \textbf{M5 -- Translation Understanding.}
The model answers questions requiring the spoken-language translation of the input SL video.
 
\end{itemize}
The proposed benchmark covers both low-level visual reasoning (M1--M3) and high-level linguistic understanding (M4--M5), providing a comprehensive evaluation of sign language understanding from multiple perspectives.

\paragraph{Generation Procedure.}
For each question category, we design a set of manually written question templates with multiple paraphrases to improve linguistic diversity. Given an SL video, we first sample a template from the corresponding category and instantiate its placeholders (e.g., glosses, positions, or gloss pairs) using the GT gloss annotations. The answer is then deterministically derived from the annotations according to the question type, ensuring that every generated QA pair is accurate and automatically verifiable. To prevent models from exploiting simple answer priors, we additionally generate a portion of negative examples. Specifically, some questions intentionally refer to non-existent glosses, out-of-range positions, or ambiguous references caused by repeated glosses, where the correct answers are ``not found,'' ``out of range,'' or ``ambiguous,'' respectively. This design encourages genuine reasoning over the sign sequence instead of memorizing frequent answer patterns. Overall, we design 30 question templates across the five question categories, with multiple paraphrases for each template in both German (PHOENIX14T) and Chinese (CSL-Daily). All random sampling is performed with fixed seeds to ensure exact reproducibility. Representative examples are provided in the Appendix.

\subsection{Baseline Model.} 
Inspired by previous SLU models~\cite{min2021visual, gan2025mixsigngraph},  as shown in Figure~\ref{fig:model}, our model for SLQA consists of three main components: a visual backbone, a QCMTD module for temporal downsampling, and a language model for answer generation. Specifically, given a sign language video $V=\{f_t\}_{t=1}^{T}$, the visual encoder first extracts frame-level features as
$E_v = \mathcal{VE}(V)$, where $E_v \in \mathbb{R}^{T \times D}$.  Then, we apply the QCMTD module to integrate question information and perform adaptive temporal downsampling:
$E_t = \mathcal{QCMTD}(E_v, Q)$, where $E_t \in \mathbb{R}^{T' \times D}$ and $T' < T$.
Finally, the downsampled question-aware video representation is fed into a language model together with the question to generate the final answer:
$A = \mathcal{LM}(E_t, Q)$.

\begin{table*}[t]
\centering
\resizebox{0.95\textwidth}{!}{
\begin{tabular}{l|cc|cc|cccccc|cccccc}
\toprule
\multirow{3}{*}{\textbf{Downsampling Module}}
&
\multicolumn{2}{c|}{\textbf{Gloss Recognition}}
&
\multicolumn{2}{c|}{\textbf{Structured QA}}
&
\multicolumn{6}{c|}{\textbf{Translation Understanding (Dev)}}
&
\multicolumn{6}{c}{\textbf{Translation Understanding (Test)}}
\\

\cmidrule(lr){2-3}
\cmidrule(lr){4-5}
\cmidrule(lr){6-11}
\cmidrule(lr){12-17}

&\multicolumn{2}{c|}{WER$\downarrow$}
& \multicolumn{2}{c|}{F1$\uparrow$} &
BLEURT$_{\text{SLT}}$ & RL & B1 & B2 & B3 & B4 &
BLEURT$_{\text{SLT}}$ & RL & B1 & B2 & B3 & B4 \\
& Dev & Test & Dev & Test &
\\

\midrule
TemConv~\cite{min2021visual}
&33.22 & 34.21
&50.66 & 49.34
&53.62 & 43.65 &35.78	&25.37 & 20.12 & 16.02
&53.03 &43.02 &35.87 &25.14  &17.53 &12.58 
\\

Q-Former~\cite{li2023blip}
&34.81   &36.92 	
&48.57   &48.86 
&46.46  &40.62	&36.74	&26.53  &20.65  &16.46 
&46.84  &37.84	&36.25	&24.35	&17.71  &13.45

 \\

\baseline{QCMTD}
&\baseline{31.67} & \baseline{33.58}
&\baseline{51.61} & \baseline{50.57}
&\baseline{55.14} & \baseline{44.91} &\baseline{37.93} &\baseline{27.61} & \baseline{21.39} & \baseline{17.33}
&\baseline{55.50} &\baseline{44.85} &\baseline{36.82} &\baseline{25.80} &\baseline{19.53} &\baseline{15.58}

\\

\bottomrule
\end{tabular}}
\caption{
Performance with different downsampling modules on Phoenix14T-QA. For gloss recognition, we report WER. For structured QA, we report F1. For translation understanding, RL denotes ROUGE-L F1 and B1--B4 denote BLEU-1 to BLEU-4.
}

\label{tab:downsampling}
\end{table*}

\begin{table*}[t]
\centering
\resizebox{0.95\textwidth}{!}{
\begin{tabular}{l|ccccccc|ccccccc}
\toprule
\multirow{2}{*}{\textbf{Downsampling Module}}
&
\multicolumn{7}{c|}{\textbf{Overall Dev}}
&
\multicolumn{7}{c}{\textbf{Overall Test}}
\\
\cmidrule(lr){2-8}
\cmidrule(lr){9-15}
&
RL & B1 & B2 & B3 & B4 & BLEURT & CIDEr
&
RL & B1 & B2 & B3 & B4 & BLEURT & CIDEr
\\
\midrule

TemConv~\cite{min2021visual}
& 52.42 &54.24 &45.42 &	38.64 &	32.64 	&54.67 	&2.742
& 52.53 &54.53 &43.64 &37.24 &33.13 &54.53 &2.642
\\

Q-Former~\cite{li2023blip}
& 53.81 &54.47 &44.25 &	38.06 &	33.58 	&55.04 	&2.843
& 53.00 &53.61 &43.26 &36.97 &32.41 &54.86 &2.722
\\

\baseline{QCMTD}
& \baseline{55.16} & \baseline{57.05} & \baseline{46.60} & \baseline{40.16} & \baseline{35.30} & \baseline{56.92} & \baseline{2.923}
& \baseline{54.16} & \baseline{56.25} & \baseline{45.58} & \baseline{39.02} & \baseline{34.12} & \baseline{56.21} & \baseline{2.768}
\\

\bottomrule
\end{tabular}}
\caption{Overall SLQA performance on Phoenix14T-QA with different downsampling modules.}
\label{tab:downsampling_all}
\end{table*}

\begin{table*}[t]
\centering
\resizebox{0.95\textwidth}{!}{
\begin{tabular}{l|cc|cc|cccccc|cccccc}
\toprule
\multirow{3}{*}{\textbf{Training Strategy}}
&
\multicolumn{2}{c|}{\textbf{Gloss Recognition}}
&
\multicolumn{2}{c|}{\textbf{Structured QA}}
&
\multicolumn{6}{c|}{\textbf{Translation Understanding (Dev)}}
&
\multicolumn{6}{c}{\textbf{Translation Understanding (Test)}}
\\

\cmidrule(lr){2-3}
\cmidrule(lr){4-5}
\cmidrule(lr){6-11}
\cmidrule(lr){12-17}

&\multicolumn{2}{c|}{WER$\downarrow$} &\multicolumn{2}{c|}{F1$\uparrow$} &
BLEURT$_{\text{SLT}}$ & RL & B1 & B2 & B3 & B4 &

BLEURT$_{\text{SLT}}$ & RL & B1 & B2 & B3 & B4
\\
&Dev & Test &Dev & Test &&&&&&&&&&&&
\\

\midrule

End-to-End
& 100.00 &	100.00
&40.84  &  40.65
&36.59 &20.9& 13.52	&8.01	&5.29	&3.87	
&36.59 &20.38& 13.58&7.89 &5.11	 &3.74

\\

+ Stage 1 (CSLR)
&34.74 &  35.62
&50.58 &   47.90
&27.75 &25.73 &14.97 &9.51  &7.23 &6.19 	
&27.45 &24.9 &14.74  &9.06  &7.03 &6.13 
\\

+ Stage 2 (SLT)
&\baseline{31.67} & \baseline{33.58}
&\baseline{51.61} & \baseline{50.57}
&\baseline{55.14} & \baseline{44.91} &\baseline{37.93} &\baseline{27.61} & \baseline{21.39} & \baseline{17.33}
&\baseline{55.50} &\baseline{44.85} &\baseline{36.82} &\baseline{25.80} &\baseline{19.53} &\baseline{15.58}
\\

\bottomrule
\end{tabular}}
\caption{
Effect of the proposed in-domain knowledge transfer strategy on Phoenix14T-QA.
Stage 1 denotes CSLR pre-training, while Stage 2 further introduces SLT training before fine-tuning on SLQA. }
\label{tab:knowledge_transfer}
\end{table*}

\begin{table*}[t]
\centering
\resizebox{0.95\textwidth}{!}{
\begin{tabular}{l|ccccccc|ccccccc}
\toprule
\multirow{2}{*}{\textbf{Training Strategy}}
&
\multicolumn{7}{c|}{\textbf{Overall Dev}}
&
\multicolumn{7}{c}{\textbf{Overall Test}}
\\
\cmidrule(lr){2-8}
\cmidrule(lr){9-15}
& RL & B1 & B2 & B3 & B4 & BLEURT & CIDEr
& RL & B1 & B2 & B3 & B4 & BLEURT & CIDEr
\\
\midrule

End-to-End
& 42.38	&29.86	&21.8	&17.3&	14.06	&43.52&	1.315
&42.42	&30.93	&22.55&	17.78	&14.37	&43.53&	1.236

\\

+ Stage 1 (CSLR)
&47.85 &	46.03 &	36.65 &	31.38 &	27.86 &	50.66 	&2.564 
&46.61 &	45.49 &	35.91 &	30.65 &	27.17 &	49.90 &	2.412\\

+ Stage 2 (SLT)
& \baseline{55.16} & \baseline{57.05} & \baseline{46.60} & \baseline{40.16} & \baseline{35.30} & \baseline{56.92} & \baseline{2.923} 
& \baseline{54.16} & \baseline{56.25} & \baseline{45.58} & \baseline{39.02} & \baseline{34.12} & \baseline{56.21} & \baseline{2.768}\\

\bottomrule
\end{tabular}}
\vspace{-2mm}
\caption{Overall SLQA performance under different training strategies on Phoenix14T-QA.}
\vspace{-2mm}
\label{tab:knowledge_transfer_all}
\end{table*}

\begin{table*}[t]
\centering
\resizebox{0.95\textwidth}{!}{
\begin{tabular}{l|cc|cc|cccccc|cccccc}
\toprule
\multirow{3}{*}{\textbf{Model}}
&
\multicolumn{2}{c|}{\textbf{Gloss Recognition}}
&
\multicolumn{2}{c|}{\textbf{Structured QA}}
&
\multicolumn{6}{c|}{\textbf{Translation Understanding (Dev)}}
&
\multicolumn{6}{c}{\textbf{Translation Understanding (Test)}}
\\

\cmidrule(lr){2-3}
\cmidrule(lr){4-5}
\cmidrule(lr){6-11}
\cmidrule(lr){12-17}

&
\multicolumn{2}{c|}{WER$\downarrow$}
&
\multicolumn{2}{c|}{F1$\uparrow$}
& BLEURT$_{\text{SLT}}$ & RL & B1 & B2 & B3 & B4 
& BLEURT$_{\text{SLT}}$ & RL & B1 & B2 & B3 & B4 \\

& Dev & Test
& Dev & Test &&&&&& &
\\

\midrule

VideoLLaMA3-2B
&100.00& 100.00
&37.91 &37.62&
32.78  & 12.04 &13.10	&5.08&	2.06&	1.03 
&32.66 & 11.96 & 12.74&5.31&2.30	&1.03
\\

Qwen3-VL-2B-Instruct
&93.73 & 94.89  
&42.75 & 42.92   & 
36.46&22.18&23.02&13.28&8.94&6.70&
36.55&20.93&23.68&13.76&9.30&7.02 
\\

InternVL3-2B
&82.42 &82.67 
&47.74 &47.38 
&40.48& 25.58 &26.32& 16.49&11.68&9.01 
&39.61&25.60&26.6& 17.18&12.36&	9.59
\\
Sign2Text2Answer 
& 32. 34& 35.64
&50.34&49.34
&{53.31} & {43.11} &{35.21} &{25.53} & {20.34} & {15.65}
&{54.76} &{43.64} &{35.42} &{24.53} &{19.55} &{14.62} \\

\baseline{SLQAM (Sign2Answer)}
&\baseline{31.67} & \baseline{33.58}
&\baseline{51.61} & \baseline{50.57}
&\baseline{55.14} & \baseline{44.91} &\baseline{37.93} &\baseline{27.61} & \baseline{21.39} & \baseline{17.33}
&\baseline{55.50} &\baseline{44.85} &\baseline{36.82} &\baseline{25.80} &\baseline{19.53} &\baseline{15.58} \\

\bottomrule
\end{tabular}} 
\caption{Comparison on Phenix14T-QA dataset. }
\vspace{-2mm}
\label{tab:compared_phoneix14t}
\end{table*}

\begin{table*}[t]
\centering
\resizebox{0.95\textwidth}{!}{
\begin{tabular}{l|ccccccc|ccccccc}
\toprule
\multirow{2}{*}{\textbf{Model}}
&
\multicolumn{7}{c|}{\textbf{Overall Dev}}
&
\multicolumn{7}{c}{\textbf{Overall Test}}
\\
\cmidrule(lr){2-8}
\cmidrule(lr){9-15}
&
RL & B1 & B2 & B3 & B4 & BLEURT & CIDEr
&
RL & B1 & B2 & B3 & B4 & BLEURT & CIDEr
\\
\midrule
VideoLLaMA3-2B
&44.67	&43.33	&31.68	&25.61	&20.89	&37.24	&1.278
&44.69	&43.19	&31.67&	25.49	&20.69	&37.45	&1.252
\\

Qwen3-VL-2B-Instruct
& 47.71	&46.72	&34.49	&28.12	&23.39	&42.01	&1.449
&47.70 &	47.87 	&35.43 &	28.85 &	23.93 &	42.02&	1.410
\\

InternVL3-2B
& 50.85	&50.39	&37.85	&31.04	&26.01&45.15&1.746
&50.93 &51.08 &38.58 &31.81 &26.83 &45.45&1.736
\\ 

Sign2Text2Answer & 
{53.34} & {55.53} & {45.76} & {39.53} & {34.04} & {54.42} & {2.433} 
& {53.54} & {54.96} & {44.01} & {38.12} & {33.24} & {54.54} & {2.343} \\

\baseline{SLQAM (Sign2Answer)} 
& \baseline{55.16} & \baseline{57.05} & \baseline{46.60} & \baseline{40.16} & \baseline{35.30} & \baseline{56.92} & \baseline{2.923} 
& \baseline{54.16} & \baseline{56.25} & \baseline{45.58} & \baseline{39.02} & \baseline{34.12} & \baseline{56.21} & \baseline{2.768} \\

\bottomrule
\end{tabular}} 
\vspace{-2mm}
\caption{Comparison of overall performance on Phenix14T-QA datasets. }
\vspace{-3mm}
\label{tab:compared_phoneix14t_all}
\end{table*}

\begin{table*}[t]
\centering
\resizebox{0.95\textwidth}{!}{
\begin{tabular}{l|cc|cc|cccccc|cccccc}
\toprule
\multirow{3}{*}{\textbf{Model}}
&
\multicolumn{2}{c|}{\textbf{Gloss Recognition}}
&
\multicolumn{2}{c|}{\textbf{Structured QA}}
&
\multicolumn{6}{c|}{\textbf{Translation Understanding (Dev)}}
&
\multicolumn{6}{c}{\textbf{Translation Understanding (Test)}}
\\

\cmidrule(lr){2-3}
\cmidrule(lr){4-5}
\cmidrule(lr){6-11}
\cmidrule(lr){12-17}

&
\multicolumn{2}{c|}{WER$\downarrow$}
&
\multicolumn{2}{c|}{F1$\uparrow$}
& BLEURT$_{\text{SLT}}$ & RL & B1 & B2 & B3 & B4 
& BLEURT$_{\text{SLT}}$ & RL & B1 & B2 & B3 & B4 \\

& Dev & Test
& Dev & Test &&&&&& &
\\

\midrule

VideoLLaMA3-2B
&100.00 &100.00
&31.78  &31.28
&20.03&13.73&12.06	&2.56	&0.56	&0.23	
 &19.97 &	13.42	 &11.78 &	2.57	 &0.65	 &0.27
\\

Qwen3-VL-2B-Instruct
&100.00  & 100.00
&33.34   & 33.37
& 25.13 & 14.97 &  15.46 & 4.30 &1.64 &0.77 
& 24.59 & 14.69 &15.08	 &4.15	&1.31 &0.6
\\

InternVL3-2B 
& 90.63 & 	90.24
&45.27  &44.7
&31.34 &	21.93	&22.55&	10.43 &	5.74 &	3.63 
&32.42 &	22.56 &	23.23&	11.25	&6.42	&4.09\\

Sign2Text2Answer
& {36.74} & {35.65} 	
& {60.54} & {60.53}   
& {56.64} & {52.42} & {50.53} & {37.51} & {27.64} & {21.51} 
& {55.80} & {52.64} & {52.56} & {37.53} & {28.97} & {21.64} \\

\baseline{SLQAM (Sign2Answer)}
& \baseline{34.19} & \baseline{34.01} 	
& \baseline{62.56} & \baseline{62.71}   
& \baseline{58.00} & \baseline{52.99} & \baseline{52.99} & \baseline{39.71} & \baseline{30.43} & \baseline{23.95} 
& \baseline{57.87} & \baseline{53.16} & \baseline{52.56} & \baseline{39.43} & \baseline{30.30} & \baseline{23.84} \\

\bottomrule
\end{tabular}}
\vspace{-1mm} 
\caption{Comparison on CSL-daily-QA dataset. }
\vspace{-2mm}
\label{tab:compared_csl}
\end{table*}

\begin{table*}[t]
\centering
\resizebox{0.95\textwidth}{!}{
\begin{tabular}{l|ccccccc|ccccccc}
\toprule
\multirow{2}{*}{\textbf{Model}}
&
\multicolumn{7}{c|}{\textbf{Overall Dev}}
&
\multicolumn{7}{c}{\textbf{Overall Test}}
\\
\cmidrule(lr){2-8}
\cmidrule(lr){9-15}
&
RL & B1 & B2 & B3 & B4 & BLEURT & CIDEr
&
RL & B1 & B2 & B3 & B4 & BLEURT & CIDEr
\\
\midrule  

VideoLLaMA3-2B 
&60.5	&59.01	&52.79	&47.58	&43.68	&38.46	&1.753 
& 60.51	&58.83	&52.64	&47.4	&43.48	&38.24	&1.735

\\

Qwen3-VL-2B-Instruct
&60.13 &	60.97	&54.22	&48.70	&44.62	&38.76&	1.715 
&60.24	&60.83	&54.14	&48.65&	44.57	&38.32	&1.699   

\\

InternVL3-2B
& 64.33 &64.53	&57.84	&52.34	&48.21	&44.66	&2.311
&64.27	&64.46	&57.69	&52.14	&47.95	&44.31	&2.277
\\

Sign2Text2Answer
& {70.61} & {70.66} & {66.12} & {60.13} & {56.93} & {54.24} & {3.647}  
& {70.75} & {70.64} & {66.03} & {60.63} & {55.03} & {54.04} & {3.754} \\

\baseline{SLQAM (Sign2Answer)} 
& \baseline{73.39} & \baseline{75.31} & \baseline{68.93} & \baseline{63.60} & \baseline{59.16} & \baseline{60.41} & \baseline{4.179}  
& \baseline{73.52} & \baseline{75.32} & \baseline{68.92} & \baseline{63.60} & \baseline{59.18} & \baseline{60.23} & \baseline{4.193} \\

\bottomrule
\end{tabular}}  
\caption{Comparison of  overall performance on CSL-daily-QA datasets. }
\vspace{-4mm}
\label{tab:compared_csl_all}
\end{table*}

\paragraph{Question-Conditioned Modulated Temporal Downsampling.}
Most existing SLT pipelines~\cite{gan2025mixsigngraph,lin2023gloss} employ a temporal 1D convolution  to  downsample SL frame features before translating the whole sentence. This design is effective for SLT, since the target text is usually semantically aligned with the entire video. However, in SLQA, the question often refers only to a specific temporal segment of the video. Directly applying question-agnostic temporal downsampling may suppress the question-relevant frames. To address this issue, we propose a \emph{QCMTD} module, which modulates video features with the question representation before each temporal pooling operation.

Specifically, as shown in Figure~\ref{fig:model}, given the input SL feature sequence
$\mathbf{V}=[\mathbf{v}_1,\mathbf{v}_2,\ldots,\mathbf{v}_T]\in\mathbb{R}^{T\times C}$
and the question representation
$\mathbf{Q}\in\mathbb{R}^{L\times D}$,
we first obtain a global question embedding $\mathbf{q}$ by average pooling over all question tokens. The visual features and the global question embedding are then projected into a shared hidden space:
\[
\mathbf{h}_t=\tanh(\mathbf{W}_v\mathbf{v}_t+\mathbf{W}_q\mathbf{q}),
\]
where $\mathbf{W}_v$ and $\mathbf{W}_q$ are learnable projection matrices.
The temporal relevance weight is computed as
\[
\alpha_t=\sigma(\mathbf{w}_a^\top\mathbf{h}_t),
\]
where $\sigma(\cdot)$ denotes the sigmoid function.  The relevance score \(\alpha_t\) indicates how important the \(t\)-th frame is with respect to the input question. Instead of directly pooling the original visual features, we use the question feature to generate modulation parameters:
\[
\boldsymbol{\gamma}_q = \mathbf{W}_{\gamma}\mathbf{q}, 
\qquad
\boldsymbol{\beta}_q = \mathbf{W}_{\beta}\mathbf{q},
\]
where \(\boldsymbol{\gamma}_q,\boldsymbol{\beta}_q \in \mathbb{R}^{D}\). The visual feature at each timestep is then modulated as:
\[
\tilde{\mathbf{v}}_t
=
\mathbf{v}_t \odot \left(1 + \alpha_t \boldsymbol{\gamma}_q\right)
+
\alpha_t \boldsymbol{\beta}_q,
\]
where \(\odot\) denotes element-wise multiplication. This formulation allows the model to enhance or suppress different visual channels according to the question, while the temporal relevance weight \(\alpha_t\) ensures that the modulation is stronger on question-relevant frames. After modulation, temporal downsampling is performed on the question-aware features:
\[
\mathbf{V}' = \operatorname{Pool}\left(
\operatorname{Conv1D}(\tilde{\mathbf{V}})
\right),
\]
where \(\tilde{\mathbf{V}}=[\tilde{\mathbf{v}}_1,\ldots,\tilde{\mathbf{v}}_T]\) denotes the modulated video features. In practice, we insert the proposed modulation before each temporal max-pooling layer in the original temporal convolution block. Therefore, the module preserves the efficiency of standard temporal  downsampling while making the downsampling process adaptive to the question. This encourages the network to retain video segments that are more informative for answering the given question.

\paragraph{In-domain Knowledge Transfer Training.}
Training an SLQA model directly on the proposed SLQA datasets often leads to unsatisfactory performance due to the limited amount of training data and the complexity of question-driven reasoning. To address this issue, we propose an \emph{in-domain knowledge transfer} strategy that progressively transfers knowledge from existing sign language understanding tasks to SLQA. Specifically, the training pipeline consists of three stages: (1) visual backbone pre-training via CSLR, (2) semantic knowledge transfer through SLT, and (3) task-specific fine-tuning on SLQA.

\noindent \textbf{Stage 1: CSLR pre-training.}
Following previous sign language understanding works, we first pre-train the visual encoder $\mathcal{VE}$ on the Continuous Sign Language Recognition (CSLR) task. Given a sign language video $f$ and its corresponding gloss sequence $g$, the visual encoder is optimized using the Connectionist Temporal Classification (CTC) objective to learn temporally aligned sign representations:
\begin{equation}
\label{equ:CTC_expectation_form}
\Theta_{\mathcal{VE}, w}^{*}
=
\mathop{\arg\min}_{\Theta_{\mathcal{VE}, w}}
\mathbb{E}_{(f,g)\sim\mathcal{D}}
\left[
\mathcal{L}_{\text{CTC}}
\left(
\mathcal{VE}(f)\cdot w,
g
\right)
\right],
\end{equation}
where $\mathcal{L}_{\text{CTC}}$ denotes the CTC loss, $w$ is the gloss classifier, and $\mathcal{D}$ is the CSLR training dataset.

\noindent \textbf{Stage 2: SLT Training.}
After obtaining a well-trained visual encoder, we use its extracted visual representations as input to a language model $\mathcal{TR}$ (\ie mT5) for SLT. The language model is optimized using the standard cross-entropy loss:
\begin{equation}
\label{equ:CEloss}
\Theta_{\mathcal{TR}}^{*}
=
\mathop{\arg\min}_{\Theta_{\mathcal{TR}}}
\mathbb{E}_{(v,t)\sim\mathcal{D}}
\left[
-\log
p_{\Theta_{\mathcal{TR}}}
\left(
t
\mid
\mathcal{TR}(v)
\right)
\right],
\end{equation}
where $v=\mathcal{VE}(f)$ denotes the visual representation extracted by the pre-trained visual encoder.

\noindent \textbf{Stage 3: SLQA fine-tuning.}
Finally, the well-trained SLT model is fine-tuned on the proposed SLQA datasets using a token-weighted cross-entropy loss. Since answers are generated from predefined templates, template tokens contribute little to SLU. Therefore, we assign smaller weights $w_i$ to template tokens and larger weights to answer-specific tokens, encouraging the model to focus on predicting the semantic content rather than memorizing fixed answer patterns.
\begin{equation}
\small
\label{equ:QAloss}
\Theta^{*} = \mathop{\arg\min}_{\Theta}
\mathbb{E}_{(v,q,a)\sim\mathcal{D}}
\left[
-\sum_{i=1}^{|a|}
w_i
\log
p_{\Theta}
\left(
a_i
\mid
a_{<i},v,q
\right)
\right]
\end{equation}

\section{Experiments}

\paragraph{Implementation Details.}
For our baseline model, we adopt ResNet-18 as the visual backbone for extracting SL features, followed by the proposed QCMTD module to aggregate and compress temporal representations. For answer generation, we employ mT5 as the language decoder. The entire model is implemented in PyTorch 2.6 and trained on two NVIDIA V100 GPUs using mixed-precision training with FP16. During training, we assign different weights to template and answer tokens in the answer-generation loss. Specifically, the weights for the template tokens and answer tokens are set to $0.02$ and $0.98$, respectively. During the SLT training stage, $\boldsymbol{\gamma}_q,\boldsymbol{\beta}_q$ in  QCMTD are set to 0.

\paragraph{Evaluation Metrics.}
Since {SLQA} covers multiple question categories, we adopt different evaluation metrics according to the answer format. For \textbf{Gloss Recognition} questions, where the answers are gloss sequences, we report the WER, following the standard evaluation protocol for CSLR. For \textbf{Translation Understanding} questions, whose answers are spoken-language sentences, we adopt ROUGE-L F1~\cite{lin2004rouge} and BLEU-1/2/3/4~\cite{papineni2002bleu}, following common practice in SLT~\cite{gan2021skeleton}. For the remaining question categories, including \textbf{Position Reasoning}, \textbf{Structural Reasoning}, and \textbf{Visual Search}, where the answers are short spans or discrete tokens, we report the F1-score to measure answer correctness. In addition, to comprehensively evaluate the semantic quality of generated answers across all question categories \textbf{(overall performance)}, we report ROUGE, BLEU, BLEURT and CIDEr. Together, these metrics provide a comprehensive evaluation of both the lexical accuracy and semantic correctness of the generated answers.

\section{Ablation Study}
\label{sec:ablation}
To better show the contribution of each component, we conduct  ablation studies on the PHOENIX14T-QA dataset.

\paragraph{Ablation on the Temporal Downsampling Module.}

We compare the proposed QCMTD module with two representative temporal aggregation strategies: the widely adopted temporal 1D convolution used in CSLR and SLT, and the Q-Former architecture commonly employed in vision-language models. As shown in Tables~\ref{tab:downsampling} and~\ref{tab:downsampling_all}, the Q-Former achieves performance comparable to temporal 1D convolution, but does not consistently improve over this simple baseline. In contrast, the QCMTD consistently outperforms both alternatives across all question categories. Moreover, QCMTD achieves the best overall performance on the \textsc{SignQA} benchmark, yielding consistent improvements in both lexical and semantic metrics (BLEURT and CIDEr). These results demonstrate that incorporating question-conditioned temporal feature aggregation enables the model to focus on question-relevant temporal regions, leading to more accurate and semantically meaningful answers.

\paragraph{Ablation on In-domain knowledge transfer training.}

Tables~\ref{tab:knowledge_transfer} and~\ref{tab:knowledge_transfer_all} evaluate the effectiveness of the proposed in-domain knowledge transfer strategy. We compare direct end-to-end training with two successive stages of in-domain pre-training: CSLR pre-training (Stage 1) and subsequent SLT pre-training (Stage 2).
Directly optimizing the model on the  SLQA objectives leads to substantially inferior performance. This is mainly because SLQA simultaneously requires fine-grained sign perception, temporal alignment, semantic understanding, and language generation, while the available QA data are insufficient to learn all these capabilities from scratch. As a result, the model struggles to establish reliable visual representations and cross-modal semantic alignment through the QA objective alone.
In contrast, in-domain knowledge transfer provides a progressive learning path. CSLR pre-training first equips the visual encoder with sign-specific representations and temporal alignment through gloss-level supervision, which is particularly beneficial for recognition and fine-grained reasoning. Building upon these visual representations, SLT pre-training further introduces sentence-level semantic alignment and language generation capabilities. This enables the model to bridge visual sign representations with natural language before being fine-tuned for the more diverse SLQA objectives. Consequently, the two stages provide complementary benefits: CSLR establishes the visual and temporal foundation, while SLT further develops cross-modal semantic understanding and generation ability. The consistent improvements across the different SLQA categories and overall evaluation metrics demonstrate the effectiveness of this progressive in-domain knowledge transfer strategy.

\section{Comparisons}   
Since SLQA is a newly proposed task and no dedicated baselines are  available, we select several representative general-purpose video understanding models and fine-tune them on the proposed {SLQA} datasets for comparison. Specifically, we compare our proposed {SLQAM} with VideoLLaMA3-2B~\cite{zhang2025videollama}, Qwen3-VL-2B-Instruct~\cite{bai2025qwen3}, and InternVL3-2B~\cite{wang2024enhancing}. \textbf{ 
A natural alternative to our end-to-end SLQAM is a cascaded pipeline that first converts the input sign video into textual representations and then performs question answering. To evaluate this alternative, we construct a two-stage baseline based on our trained gloss-based sign language translation (GBSLT) model. Specifically, the GBSLT model first predicts both the gloss sequence and the spoken-language translation of the input video. The predicted gloss sequence and translation, together with the input question, are then fed into a separately trained mT5 model used exclusively for question answering. We denote this cascaded baseline as \textbf{Sign2Text2Answer}, while our end-to-end SLQAM, which directly generates the answer from the sign video and question, is denoted as \textbf{Sign2Answer}.}

\paragraph{Evaluation of SLQA Performance on PHOENIX14T-QA.}

Tables~\ref{tab:compared_phoneix14t} and~\ref{tab:compared_phoneix14t_all} compare SLQAM with general-purpose VLMs and the cascaded {Sign2Text2Answer} baseline on PHOENIX14T-QA. General-purpose VLMs perform poorly, particularly on gloss recognition and structured QA, indicating that general video-language pre-training is insufficient to capture the fine-grained hand movements and temporal structures of sign language. By first predicting gloss and translation sequences and then performing text-based QA, Sign2Text2Answer substantially outperforms the general-purpose VLMs, demonstrating the importance of in-domain CSLR and SLT knowledge. Nevertheless, SLQAM consistently achieves the best performance across gloss recognition, structured QA, translation understanding, and overall metrics. This suggests that directly conditioning visual representations on the question is more effective than relying on explicitly predicted intermediate sequences, which may introduce error propagation in the cascaded pipeline.

\paragraph{Evaluation of SLQA Performance on CSL-Daily-QA.}

Tables~\ref{tab:compared_csl} and~\ref{tab:compared_csl_all} compare SLQAM with general-purpose VLMs and the cascaded {Sign2Text2Answer} baseline on CSL-Daily-QA da. Consistent with the results on PHOENIX14T-QA, general-purpose VLMs perform poorly, particularly on gloss recognition and structured QA, indicating their limited ability to capture fine-grained sign articulations and temporal relationships. Sign2Text2Answer substantially outperforms VLMs by leveraging predicted gloss and translation sequences, demonstrating the importance of in-domain sign-language representations. Nevertheless, SLQAM achieves the best overall performance and consistently higher results on structured QA and translation understanding. These results suggest that direct question-conditioned video modeling can better preserve question-relevant visual information and mitigate the error propagation caused by explicitly predicted intermediate representations.

\section{Conclusion}

In this paper, we introduced a new SL understanding task, \emph{Sign Language Question Answering} (SLQA), which aims to evaluate sign language understanding through natural language question answering rather than fixed prediction objectives. To facilitate this task, we constructed two {SLQA} benchmarks based on the widely used PHOENIX14T and CSL-Daily datasets by leveraging their existing gloss and translation annotations. The resulting benchmarks cover five complementary question categories, including \emph{Position Reasoning}, \emph{Structural Reasoning}, \emph{Visual Search}, \emph{Gloss Recognition}, and \emph{Translation Understanding}, providing a more comprehensive evaluation of SLU. Furthermore, we proposed a simple yet effective baseline model, \textsc{SLQAM}, equipped with two key components: (1) an in-domain knowledge transfer strategy that progressively transfers knowledge from CSLR and SLT to SLQA, and (2) a QCMTD module that dynamically selects question-relevant temporal features for answer generation. Extensive experiments on both benchmarks demonstrate that the proposed baseline consistently outperforms existing general-purpose video question answering models, highlighting the unique challenges of SLQA and the effectiveness of the proposed framework. We hope that {SLQA} will serve as a standardized benchmark for evaluating SLU beyond conventional recognition and translation tasks, and inspire future research on more general and reasoning-oriented SL foundation models.

\bibliography{bib/cslr, bib/islr, bib/other, bib/slt, bib/signqa, bib/videoQA}

@inproceedings{min2021visual,
  title={Visual alignment constraint for continuous sign language recognition},
  author={Min, Yuecong and Hao, Aiming and Chai, Xiujuan and Chen, Xilin},
  booktitle={ICCV},
  pages={11542--11551},
  year={2021}
}

@inproceedings{gan2024signgraph,
  title={SignGraph: A Sign Sequence is Worth Graphs of Nodes},
  author={Gan, Shiwei and Yin, Yafeng and Jiang, Zhiwei and Wen, Hongkai and Xie, Lei and Lu, Sanglu},
  booktitle={Proceedings of the IEEE/CVF Conference on Computer Vision and Pattern Recognition},
  pages={13470--13479},
  year={2024}
}

@inproceedings{min2025closer,
  title={A Closer Look at Skeleton-based Continuous Sign Language Recognition},
  author={Min, Yuecong and Yang, Yifan and Jiao, Peiqi and Nan, Zixi and Chen, Xilin},
  booktitle={Proceedings of the IEEE/CVF International Conference on Computer Vision},
  pages={4909--4915},
  year={2025}
}

@inproceedings{hu2021hand,
	title={Hand-Model-Aware Sign Language Recognition},
	author={Hu, Hezhen and Zhou, Wengang and Li, Houqiang},
	booktitle={Proceedings of the AAAI Conference on Artificial Intelligence},
	volume={35},

	pages={1558--1566},
	year={2021}
}

@article{zhou2025scaling,
  title={Scaling up multimodal pre-training for sign language understanding},
  author={Zhou, Wengang and Zhao, Weichao and Hu, Hezhen and Li, Zecheng and Li, Houqiang},
  journal={TPAMI},
  year={2025},
  publisher={IEEE}
}

@inproceedings{zuo2023natural,
  title={Natural language-assisted sign language recognition},
  author={Zuo, Ronglai and Wei, Fangyun and Mak, Brian},
  booktitle={Proceedings of the IEEE/CVF conference on computer vision and pattern recognition},
  pages={14890--14900},
  year={2023}
}

@article{wong2025signrep,
  title={Signrep: Enhancing self-supervised sign representations},
  author={Wong, Ryan and Camgoz, Necati Cihan and Bowden, Richard},
  journal={arXiv preprint arXiv:2503.08529},
  year={2025}
}

@inproceedings{papineni2002bleu,
	title={BLEU: a method for automatic evaluation of machine translation},
	author={Papineni, Kishore and Roukos, Salim and Ward, Todd and Zhu, Wei-Jing},
	booktitle={ACL},
	pages={311--318},
	year={2002}
}

@inproceedings{lin2004rouge,
	title={Rouge: A package for automatic evaluation of summaries},
	author={Lin, Chin-Yew},
	booktitle={Text summarization branches out},
	pages={74--81},
	year={2004}
}

@article{liu2026ssl,
  title={SSL-SSAW: Self-Supervised Learning with Sigmoid Self-Attention Weighting for Question-Based Sign Language Translation},
  author={Liu, Zekang and Feng, Wei and Shang, Fanhua and Hu, Lianyu and Feng, Jichao and Gao, Liqing},
  journal={Pattern Recognition},
  pages={114189},
  year={2026},
  publisher={Elsevier}
}

@article{gao2024overcoming,
  title={Overcoming modality bias in question-driven sign language video translation},
  author={Gao, Liqing and Lyu, Fan and Shi, Peng and Zhu, Lei and Pu, Junfu and Wan, Liang and Feng, Wei},
  journal={IEEE Transactions on Circuits and Systems for Video Technology},
  volume={34},
  number={11},
  pages={11724--11738},
  year={2024},
  publisher={IEEE}
}

@article{testa2026slu,
  title={SLU-2K: A Question-Based Benchmark for Semantic Evaluation of Sign Language Translation},
  author={Testa, Zeno and Furnari, Antonino and Baraldi, Lorenzo and D{\'\i}az-Rodr{\'\i}guez, Natalia},
  journal={arXiv preprint arXiv:2606.03788},
  year={2026}
}

@inproceedings{gan2021skeleton,
  title={Skeleton-Aware Neural Sign Language Translation},
  author={Gan, Shiwei and Yin, Yafeng and Jiang, Zhiwei and Xie, Lei and Lu, Sanglu},
  booktitle={MM},
  pages={4353--4361},
  year={2021}
}

@article{chen2022two,
  title={Two-stream network for sign language recognition and translation},
  author={Chen, Yutong and Zuo, Ronglai and Wei, Fangyun and Wu, Yu and Liu, Shujie and Mak, Brian},
  journal={Advances in Neural Information Processing Systems},
  volume={35},
  pages={17043--17056},
  year={2022}
}

@inproceedings{cihan2018neural,
	title={Neural sign language translation},
	author={Camgoz, Necati Cihan and Hadfield, Simon and Koller, Oscar and Ney, Hermann and Bowden, Richard},
	booktitle={CVPR},
	pages={7784--7793},
	year={2018}
}

@inproceedings{zhou2021improving,
	title={Improving Sign Language Translation with Monolingual Data by Sign Back-Translation},
	author={Zhou, Hao and Zhou, Wengang and Qi, Weizhen and Pu, Junfu and Li, Houqiang},
	booktitle={CVPR},
	pages={1316--1325},
	year={2021}
}

@inproceedings{zhou2023gloss,
  title={Gloss-free sign language translation: Improving from visual-language pretraining},
  author={Zhou, Benjia and Chen, Zhigang and Clap{\'e}s, Albert and Wan, Jun and Liang, Yanyan and Escalera, Sergio and Lei, Zhen and Zhang, Du},
  booktitle={ICCV},
  pages={20871--20881},
  year={2023}
}

@misc{gan2025mixsigngraph,
    title={MixSignGraph: A Sign Sequence is Worth Mixed Graphs of Nodes},
    author={Shiwei Gan and Yafeng Yin and Zhiwei Jiang and  Lei Xie and Sanglu Lu and Hongkai Wen},
    year={2025}, 
    booktitle={NeurIPS},
   
}

@article{liang2024llava,
  title={LLaVA-SLT: Visual Language Tuning for Sign Language Translation},
  author={Liang, Han and Huang, Chengyu and Xu, Yuecheng and Tang, Cheng and Ye, Weicai and Zhang, Juze and Chen, Xin and Yu, Jingyi and Xu, Lan},
  journal={arXiv preprint arXiv:2412.16524},
  year={2024}
}

@article{uthus2023youtube,
  title={Youtube-asl: A large-scale, open-domain american sign language-english parallel corpus},
  author={Uthus, Dave and Tanzer, Garrett and Georg, Manfred},
  journal={Advances in Neural Information Processing Systems},
  volume={36},
  pages={29029--29047},
  year={2023}
}

@article{rust2024towards,
  title={Towards privacy-aware sign language translation at scale},
  author={Rust, Phillip and Shi, Bowen and Wang, Skyler and Camg{\"o}z, Necati Cihan and Maillard, Jean},
  journal={arXiv preprint arXiv:2402.09611},
  year={2024}
}

@inproceedings{chen2024factorized,
  title={Factorized Learning Assisted with Large Language Model for Gloss-free Sign Language Translation},
  author={Chen, Zhigang and Zhou, Benjia and Li, Jun and Wan, Jun and Lei, Zhen and Jiang, Ning and Lu, Quan and Zhao, Guoqing},
  booktitle={Proceedings of the 2024 Joint International Conference on Computational Linguistics, Language Resources and Evaluation (LREC-COLING 2024)},
  pages={7071--7081},
  year={2024}
}

@article{gueuwou2025signmusketeers,
      title={Signmusketeers: An efficient multi-stream approach for sign language translation at scale},
      author={Gueuwou, Shester and Du, Xiaodan and Shakhnarovich, Greg and Livescu, Karen},
      journal={Findings of the Association for Computational Linguistics: ACL 2025},
      year={2025}
}

@inproceedings{jiang2024signclip,
  title={SignCLIP: Connecting Text and Sign Language by Contrastive Learning},
  author={Jiang, Zifan and Sant, Gerard and Moryossef, Amit and M{\"u}ller, Mathias and Sennrich, Rico and Ebling, Sarah},
  booktitle={Proceedings of the 2024 Conference on Empirical Methods in Natural Language Processing},
  pages={9171--9193},
  year={2024}
}

@inproceedings{gueuwouetal2025shubert,
    title = "{SH}u{BERT}: Self-Supervised Sign Language Representation Learning via Multi-Stream Cluster Prediction",
    author = "Gueuwou, Shester  and
      Du, Xiaodan  and
      Shakhnarovich, Greg  and
      Livescu, Karen  and
      Liu, Alexander H.", 
    booktitle = "Proceedings of the 63rd Annual Meeting of the Association for Computational Linguistics (Volume 1: Long Papers)",
    month = jul,
    year = "2025",
    address = "Vienna, Austria",
    publisher = "Association for Computational Linguistics",
    url = "https://aclanthology.org/2025.acl-long.1397/",
    pages = "28792--28810",
    ISBN = "979-8-89176-251-0", 
}

@article{li2025uni,
  title={Uni-sign: Toward unified sign language understanding at scale},
  author={Li, Zecheng and Zhou, Wengang and Zhao, Weichao and Wu, Kepeng and Hu, Hezhen and Li, Houqiang},
  journal={arXiv preprint arXiv:2501.15187},
  year={2025}
}

@inproceedings{lin2023gloss,
  title={Gloss-Free End-to-End Sign Language Translation},
  author={Lin, Kezhou and Wang, Xiaohan and Zhu, Linchao and Sun, Ke and Yang, Yi and others},
  booktitle={The 61st Annual Meeting Of The Association For Computational Linguistics},
  year={2023}
}

@inproceedings{jiao2024visual,
  title={Visual alignment pre-training for sign language translation},
  author={Jiao, Peiqi and Min, Yuecong and Chen, Xilin},
  booktitle={European Conference on Computer Vision},
  pages={349--367},
  year={2024},
  organization={Springer}
}

@article{guo2025bridging,
  title={Bridging Sign and Spoken Languages: Pseudo Gloss Generation for Sign Language Translation},
  author={Guo, Jianyuan and Li, Peike and Cohn, Trevor},
  journal={arXiv preprint arXiv:2505.15438},
  year={2025}
}

@inproceedings{wang2025gloss,
  title={Gloss Matters: Unlocking the Potential of Non-Autoregressive Sign Language Translation},
  author={Wang, Zhihao and Liu, Shiyu and He, Zhiwei and Zheng, Kangjie and Shao, Liangying and Yao, Junfeng and Su, Jinsong},
  booktitle={Proceedings of the 33rd ACM International Conference on Multimedia},
  pages={4127--4136},
  year={2025}
}

@inproceedings{zhao2024conditional,
  title={Conditional variational autoencoder for sign language translation with cross-modal alignment},
  author={Zhao, Rui and Zhang, Liang and Fu, Biao and Hu, Cong and Su, Jinsong and Chen, Yidong},
  booktitle={Proceedings of the aaai conference on artificial intelligence},
  volume={38},
  number={17},
  pages={19643--19651},
  year={2024}
}

@article{fish2025geo,
  title={Geo-Sign: Hyperbolic Contrastive Regularisation for Geometrically Aware Sign Language Translation},
  author={Fish, Edward and Bowden, Richard},
  journal={arXiv preprint arXiv:2506.00129},
  year={2025}
}

@inproceedings{liu2025scope,
  title={SCOPE: Sign Language Contextual Processing with Embedding from LLMs},
  author={Liu, Yuqi and Zhang, Wenqian and Ren, Sihan and Huang, Chengyu and Yu, Jingyi and Xu, Lan},
  booktitle={Proceedings of the AAAI Conference on Artificial Intelligence},
  volume={39},
  number={6},
  pages={5739--5747},
  year={2025}
}

@article{tanzer2024youtube,
  title={YouTube-SL-25: A Large-Scale, Open-Domain Multilingual Sign Language Parallel Corpus},
  author={Tanzer, Garrett and Zhang, Biao},
  journal={arXiv preprint arXiv:2407.11144},
  year={2024}
}

@inproceedings{rusttowards,
    title = "Towards Privacy-Aware Sign Language Translation at Scale",
    author = "Rust, Phillip and Shi, Bowen and Wang, Skyler and Camgoz, Necati Cihan and Maillard, Jean",
    booktitle = "Proceedings of the 62nd Annual Meeting of the Association for Computational Linguistics (Volume 1: Long Papers)",
    year = "2024",
    address = "Bangkok, Thailand",
    publisher = "Association for Computational Linguistics",
    url = "https://aclanthology.org/2024.acl-long.467",
    pages = "8624--8641",
}

@inproceedings{gan2026learning,
  title={Learning Effective Sign Features without Text for Gloss-free Sign Language Translation},
  author={Gan, Shiwei and Liu, Xiao and Yin, Yafeng and Liu, Nan and Liu, Kuizhuang and Tuerdaken, Desibieer and Jiang, Zhiwei and Xie, Lei and Lu, Sanglu and Wen, Hongkai},
  booktitle={Proceedings of the IEEE/CVF Conference on Computer Vision and Pattern Recognition},
  pages={9827--9836},
  year={2026}
}

@inproceedings{gan2026signllama, 
    title = {Enhancing Gloss-free Sign Language Translation by Prioritizing Visual Features for LLMs} ,
   author={Gan, Shiwei and Liu, Xiao and Yin, Yafeng and  Jiang, Zhiwei and Guo, Bowen and Xie, Lei and Lu, Sanglu and Wen, Hongkai}, 
   booktitle={Proceedings of the 31st ACM International Conference on Multimedia}, 
   year={2026}
}

@inproceedings{zhong2022video,
  title={Video question answering: Datasets, algorithms and challenges},
  author={Zhong, Yaoyao and Ji, Wei and Xiao, Junbin and Li, Yicong and Deng, Weihong and Chua, Tat-Seng},
  booktitle={Proceedings of the 2022 conference on empirical methods in natural language processing},
  pages={6439--6455},
  year={2022}
}

@inproceedings{wu2025videoqa,
  title={VideoQA-TA: Temporal-Aware Multi-Modal Video Question Answering},
  author={Wu, Zhixuan and Cheng, Bo and Han, Jiale and Ma, Jiabao and Zhang, Shuhao and Chen, Yuli and Li, Changbo},
  booktitle={Proceedings of the 31st International Conference on Computational Linguistics},
  pages={7239--7252},
  year={2025}
}

@article{ataallah2024minigpt4,
  title={MiniGPT4-Video: Advancing Multimodal LLMs for Video Understanding with Interleaved Visual-Textual Tokens},
  author={Ataallah, Kirolos and Shen, Xiaoqian and Abdelrahman, Eslam and Sleiman, Essam and Zhu, Deyao and Ding, Jian and Elhoseiny, Mohamed},
  journal={arXiv preprint arXiv:2404.03413},
  year={2024}
}

@inproceedings{li2024llama,
  title={Llama-vid: An image is worth 2 tokens in large language models},
  author={Li, Yanwei and Wang, Chengyao and Jia, Jiaya},
  booktitle={European Conference on Computer Vision},
  pages={323--340},
  year={2024},
  organization={Springer}
}

@inproceedings{antol2015vqa,
  title={Vqa: Visual question answering},
  author={Antol, Stanislaw and Agrawal, Aishwarya and Lu, Jiasen and Mitchell, Margaret and Batra, Dhruv and Zitnick, C Lawrence and Parikh, Devi},
  booktitle={Proceedings of the IEEE international conference on computer vision},
  pages={2425--2433},
  year={2015}
}

@inproceedings{johnson2017clevr,
  title={Clevr: A diagnostic dataset for compositional language and elementary visual reasoning},
  author={Johnson, Justin and Hariharan, Bharath and Van Der Maaten, Laurens and Fei-Fei, Li and Lawrence Zitnick, C and Girshick, Ross},
  booktitle={Proceedings of the IEEE conference on computer vision and pattern recognition},
  pages={2901--2910},
  year={2017}
}

@article{lu2019vilbert,
  title={Vilbert: Pretraining task-agnostic visiolinguistic representations for vision-and-language tasks},
  author={Lu, Jiasen and Batra, Dhruv and Parikh, Devi and Lee, Stefan},
  journal={Advances in neural information processing systems},
  volume={32},
  year={2019}
}

@inproceedings{li2022blip,
  title={Blip: Bootstrapping language-image pre-training for unified vision-language understanding and generation},
  author={Li, Junnan and Li, Dongxu and Xiong, Caiming and Hoi, Steven},
  booktitle={International conference on machine learning},
  pages={12888--12900},
  year={2022},
  organization={PMLR}
}

@article{alayrac2022flamingo,
  title={Flamingo: a visual language model for few-shot learning},
  author={Alayrac, Jean-Baptiste and Donahue, Jeff and Luc, Pauline and Miech, Antoine and Barr, Iain and Hasson, Yana and Lenc, Karel and Mensch, Arthur and Millican, Katherine and Reynolds, Malcolm and others},
  journal={Advances in neural information processing systems},
  volume={35},
  pages={23716--23736},
  year={2022}
}

@article{liu2023visual,
  title={Visual instruction tuning},
  author={Liu, Haotian and Li, Chunyuan and Wu, Qingyang and Lee, Yong Jae},
  journal={Advances in neural information processing systems},
  volume={36},
  pages={34892--34916},
  year={2023}
}

@inproceedings{xu2017video,
  title={Video question answering via gradually refined attention over appearance and motion},
  author={Xu, Dejing and Zhao, Zhou and Xiao, Jun and Wu, Fei and Zhang, Hanwang and He, Xiangnan and Zhuang, Yueting},
  booktitle={Proceedings of the 25th ACM international conference on Multimedia},
  pages={1645--1653},
  year={2017}
}

@inproceedings{yu2019activitynet,
  title={Activitynet-qa: A dataset for understanding complex web videos via question answering},
  author={Yu, Zhou and Xu, Dejing and Yu, Jun and Yu, Ting and Zhao, Zhou and Zhuang, Yueting and Tao, Dacheng},
  booktitle={Proceedings of the AAAI conference on artificial intelligence},
  volume={33},
  number={01},
  pages={9127--9134},
  year={2019}
}

@inproceedings{lin2024video,
  title={Video-llava: Learning united visual representation by alignment before projection},
  author={Lin, Bin and Ye, Yang and Zhu, Bin and Cui, Jiaxi and Ning, Munan and Jin, Peng and Yuan, Li},
  booktitle={Proceedings of the 2024 conference on empirical methods in natural language processing},
  pages={5971--5984},
  year={2024}
}

@inproceedings{maaz2024video,
  title={Video-chatgpt: Towards detailed video understanding via large vision and language models},
  author={Maaz, Muhammad and Rasheed, Hanoona and Khan, Salman and Khan, Fahad},
  booktitle={Proceedings of the 62nd Annual Meeting of the Association for Computational Linguistics (Volume 1: Long Papers)},
  pages={12585--12602},
  year={2024}
}

@inproceedings{li2023blip,
  title={Blip-2: Bootstrapping language-image pre-training with frozen image encoders and large language models},
  author={Li, Junnan and Li, Dongxu and Savarese, Silvio and Hoi, Steven},
  booktitle={International conference on machine learning},
  pages={19730--19742},
  year={2023},
  organization={PMLR}
}

@article{zhang2025videollama,
  title={Videollama 3: Frontier multimodal foundation models for image and video understanding},
  author={Zhang, Boqiang and Li, Kehan and Cheng, Zesen and Hu, Zhiqiang and Yuan, Yuqian and Chen, Guanzheng and Leng, Sicong and Jiang, Yuming and Zhang, Hang and Li, Xin and others},
  journal={arXiv preprint arXiv:2501.13106},
  year={2025}
}

@article{bai2025qwen3,
  title={Qwen3-vl technical report},
  author={Bai, Shuai and Cai, Yuxuan and Chen, Ruizhe and Chen, Keqin and Chen, Xionghui and Cheng, Zesen and Deng, Lianghao and Ding, Wei and Gao, Chang and Ge, Chunjiang and others},
  journal={arXiv preprint arXiv:2511.21631},
  year={2025}
}

@article{wang2024enhancing,
  title={Enhancing the reasoning ability of multimodal large language models via mixed preference optimization},
  author={Wang, Weiyun and Chen, Zhe and Wang, Wenhai and Cao, Yue and Liu, Yangzhou and Gao, Zhangwei and Zhu, Jinguo and Zhu, Xizhou and Lu, Lewei and Qiao, Yu and others},
  journal={arXiv preprint arXiv:2411.10442},
  year={2024}
}


\clearpage
\newpage

\begin{table*}[t]
\centering
\small
\setlength{\tabcolsep}{7pt}
\renewcommand{\arraystretch}{1.15}
\begin{tabular}{lcccccccc}
\toprule
\multirow{2}{*}{\textbf{Question Category}} &
\multirow{2}{*}{\textbf{\#Templates}} &
\multicolumn{3}{c}{\textbf{PHOENIX-2014T}} &
\multicolumn{3}{c}{\textbf{CSL-Daily}} \\
\cmidrule(lr){3-5}
\cmidrule(lr){6-8}
&
& Train & Dev & Test & Train & Dev & Test \\
\midrule

M1 Position Reasoning                & 9      & 7,096 & 519 & 642 & 18,400 & 1,077 & 1,176 \\

M2 Structural Reasoning              & 13     & 7,096 & 519 & 642 & 18,400 & 1,077 & 1,176 \\

M3 Visual Search$^\dagger$           & 5(+1)      & 7,096 & 519 & 642 & 18,400 & 1,077 & 1,176 \\

M4 Gloss Recognition                 & 1      & 7,096 & 519 & 642 & 18,400 & 1,077 & 1,176 \\

M5 Translation Understanding         & 1      & 7,096 & 519 & 642 & 18,400 & 1,077 & 1,176 \\

\midrule
\textbf{Total QA Pairs} & \textbf{30} &
35,480 & 2,595 & 3,210 &
92,000 & 5,385 & 5,880 \\

\bottomrule
\end{tabular}

\caption{{SLQA} datasets statistics. One question per module per video; splits
follow the official corpus partitions (train/dev/test:
$35{,}480$/$2{,}595$/$3{,}210$ for PHOENIX-2014T and
$92{,}000$/$5{,}385$/$5{,}880$ for CSL-Daily). $^\dagger$M3 includes one CSL-only
non-manual (interrogative) template ($3{,}448$ QA); PHOENIX-2014T is
declarative-only.}

\label{tab:signqa_stats}
\end{table*}

\begin{CJK}{UTF8}{gbsn}

\begin{table*}[t]
\centering
\footnotesize
\setlength{\tabcolsep}{6pt}
\renewcommand{\arraystretch}{1.2}

\begin{tabular}{@{}l p{\dimexpr\textwidth-4em-12pt\relax}@{}}
\toprule
\textbf{Module} & \textbf{Representative Question--Answer Pair} \quad\textit{(English translation)} \\
\midrule

M1 &
\textbf{[de]} Welches Gloss steht an Position~2?
$\rightarrow$ ``ZUSCHAUER''. \\ 
& \textit{(Which gloss appears at position~2? $\rightarrow$ \textsc{viewer}.)}
\\

\addlinespace

M2 &
\textbf{[zh]} 第一个动作之后的那个是什么？
$\rightarrow$ 第2个 gloss 为``时间''。 \\
& \textit{(Which gloss comes immediately after the first sign? $\rightarrow$ the second gloss is \textsc{time}.)}
\\

\addlinespace

M3$_{\text{man.}}$ &
\textbf{[de]} Erstes Vorkommen von ``KUESTE''?
$\rightarrow$ Position~6. \\
& \textit{(Where does \textsc{coast} first appear? $\rightarrow$ position~6.)}
\\

\addlinespace

M3$_{\text{n.m.}}$ &
\textbf{[zh]} 结合面部表情，这是疑问句吗？
$\rightarrow$ 不是疑问句。 \\ 
&\textit{(Based on the non-manual markers, is this an interrogative sentence? $\rightarrow$ No.)}
\\

\addlinespace

M4 &
\textbf{[de]} Gloss-Sequenz des Videos?
$\rightarrow$ ``TEIL BISSCHEN FROST NACHT $\ldots$''. \\
& \textit{(What is the gloss sequence of the video? $\rightarrow$ \textsc{part bit frost night} $\ldots$)}
\\

\addlinespace

M5 &
\textbf{[zh]} 完整句子是什么？
$\rightarrow$ ``这些经典的文学作品他全都读过''。 \\
&\textit{(What is the complete spoken-language sentence? $\rightarrow$ ``He has read all these classic works.'')}
\\

\bottomrule
\end{tabular}
\caption{Representative question--answer pairs for each question category
in {SLQA}. [de] denotes PHOENIX-2014T, and [zh] denotes CSL-Daily.
Long answers are truncated for brevity. M3 includes both manual gloss
localization (M3$_{\text{man.}}$) and non-manual interrogativity detection
(M3$_{\text{n.m.}}$).}

\label{tab:signqa_examples}
\end{table*}

\end{CJK}

\begin{figure*}
    \centering
    \includegraphics[width=0.75\linewidth]{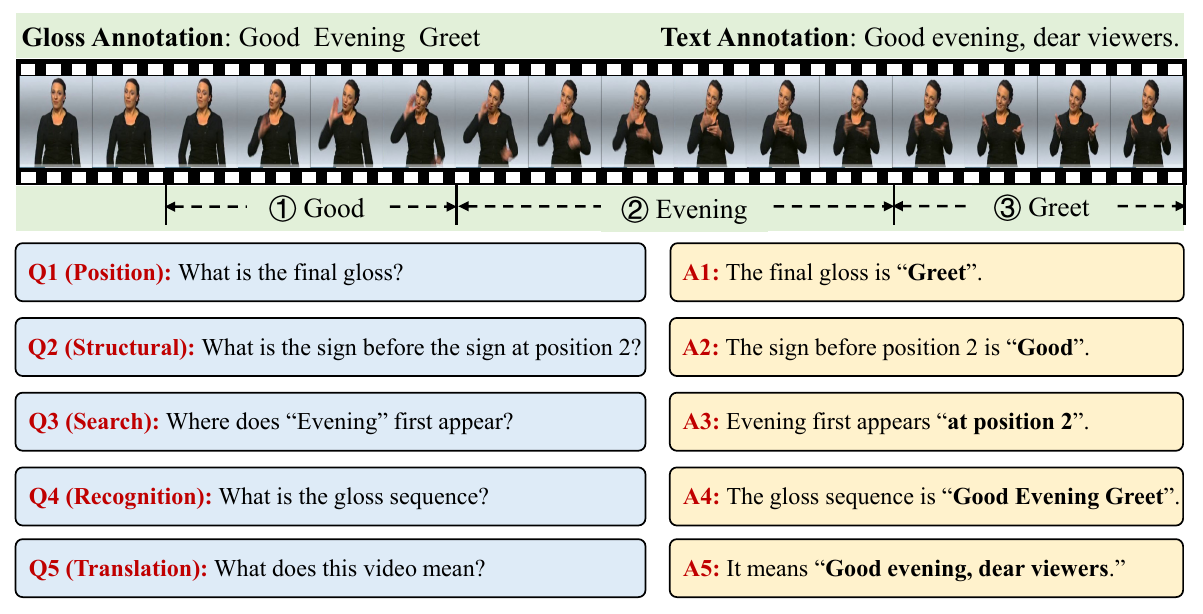}
    \caption{Illustration of each question category in SLQA.}
    \label{fig:SignQA_Motivation}
\end{figure*}

\label{sec:experimental_setting} 

\section{Experimental Details}

\paragraph{Datasets.}
We evaluate the proposed method on the \textsc{SignQA} benchmark, which is constructed from two widely used sign language translation datasets: \textbf{PHOENIX-2014T}~\cite{cihan2018neural} and \textbf{CSL-Daily}~\cite{zhou2021improving}. PHOENIX-2014T is a German Sign Language (DGS) dataset containing 7,096 training, 519 validation, and 642 test videos collected from nine signers. CSL-Daily is a Chinese Sign Language dataset consisting of 18,400 training, 1,077 validation, and 1,176 test videos from ten signers.

Following the dataset construction procedure described in Section \textbf{Dataset Construction}, we generate one question for each of the five question categories from every video, resulting in five question--answer pairs per sample. We strictly follow the official train/validation/test splits of the original datasets, ensuring that no additional data leakage is introduced.
In total, \textsc{SignQA} contains 41,285 QA pairs for PHOENIX-2014T (35,480/2,595/3,210 for train/validation/test) and 103,265 QA pairs for CSL-Daily (92,000/5,385/5,880 for train/validation/test), yielding a total of 144,550 question--answer pairs. The benchmark covers five complementary question categories, namely \emph{Position Reasoning}, \emph{Structural Reasoning}, \emph{Visual Search}, \emph{Translation Understanding}, and \emph{Gloss Recognition}. Detailed statistics are reported in Table~\ref{tab:signqa_stats}.

\paragraph{Data Preprocessing.}  We adopt following augmentations, including resizing to 256$\times$256 pixels, random cropping to 224$\times$224 pixels, random horizontal flipping (probability 0.5), and random temporal scaling within ±20\%. During inference, frames are resized to 256$\times$256 pixels and center-cropped to 224$\times$224 pixels.

\section{Examples of the Constructed SLQA Datasets}

Table~\ref{tab:signqa_examples} and Figure~\ref{fig:SignQA_Motivation} present representative question--answer
examples from the constructed \textsc{SignQA} datasets. Unlike previous
question-guided SLT tasks, where questions primarily serve as auxiliary
conditions for generating different translations, our questions directly
probe the content of the sign language video. In particular, \textsc{SignQA}
covers five complementary types of questions:
\emph{Position Reasoning}, \emph{Structural Reasoning}, \emph{Visual Search},
\emph{Gloss Recognition}, and \emph{Translation Understanding}.

As illustrated in the table, these question types require models to operate at
different levels of sign language understanding. \emph{Position Reasoning}
and \emph{Structural Reasoning} evaluate whether a model can identify and
reason over the ordered relationships among signs. \emph{Visual Search}
requires the model to locate specific signs in the video and, where
applicable, interpret non-manual markers such as facial expressions.
\emph{Gloss Recognition} evaluates fine-grained sign-level perception, while
\emph{Translation Understanding} assesses the ability to infer the overall
spoken-language meaning conveyed by the video. Together, these tasks cover
both fine-grained visual and temporal understanding and high-level semantic
interpretation, providing a more comprehensive evaluation of sign language
understanding than translation-only objectives.

\section{Ablation Study on CSL-daily-QA}

\begin{table*}[t]
\centering
\resizebox{0.98\textwidth}{!}{
\begin{tabular}{l|cc|cc|cc|cccccc|cccccc}
\toprule
\multirow{3}{*}{\textbf{Training Strategy}}
&
\multicolumn{2}{c|}{\textbf{Gloss Recognition}}
&
\multicolumn{2}{c|}{\textbf{Structured QA}}
&
\multicolumn{2}{c|}{\textbf{Non Manual}}
&
\multicolumn{6}{c|}{\textbf{Translation Understanding (Dev)}}
&
\multicolumn{6}{c}{\textbf{Translation Understanding (Test)}}
\\

\cmidrule(lr){2-19}

&
\multicolumn{2}{c|}{WER$\downarrow$}
&
\multicolumn{2}{c|}{ACC$\uparrow$}
&
\multicolumn{2}{c|}{F1$\uparrow$}
&
BLEURT$_{\text{SLT}}$ & RL & B1 & B2 & B3 & B4 &
BLEURT$_{\text{SLT}}$ & RL & B1 & B2 & B3 & B4\\
& Dev & Test & Dev & Test & Dev & Test
& & & & & & & & & & & &\\
\midrule
 
+ Stage 1 (CSLR) 
& 34.17 &	33.67 
&64.23 &	64.36
& 90.43  &92.15 
&57.56 	&53.27	&52.75	&39.33 	&29.86 &	23.21 
&58.32 &	53.67&	52.74	&39.54 &	30.27 &	23.67

\\

+ Stage 2 (SLT) 
& 34.19 &34.01 	
&62.56 &62.71 
& 91.49 &93.19 
&58.00 &52.99	&52.99	&39.71 	&30.43 	&23.95 
&57.87 &	53.16&	52.56	&39.43& 	30.30 &	23.84 
\\
\bottomrule
\end{tabular}}
\caption{Effect of the proposed in-domain knowledge transfer strategy on CSL-daily-QA.}
\label{tab:knowledge_transfer_CSL}
\end{table*}

\begin{table*}[t]
\centering
\resizebox{0.98\textwidth}{!}{
\begin{tabular}{l|ccccccc|ccccccc}
\toprule
\multirow{2}{*}{\textbf{Training Strategy}}
&
\multicolumn{7}{c|}{\textbf{Overall Dev}}
&
\multicolumn{7}{c}{\textbf{Overall Test}}
\\
\cmidrule(lr){2-8}
\cmidrule(lr){9-15}
& RL & B1 & B2 & B3 & B4 & BLEURT & CIDEr
& RL & B1 & B2 & B3 & B4 & BLEURT & CIDEr
\\
\midrule


+ Stage 1 (CSLR)
&73.07 &	74.76 &	68.35 &	62.98 &	58.81 	&60.27 	&4.150
&73.46 &	75.20 &	68.79 	&63.44 &	59.02 	&60.51& 	4.212
 \\

+ Stage 2 (SLT) 
&73.39 	&75.31 &68.93 	&63.60 	&59.16 	&60.41 &4.179  
&73.52 	&75.32 	&68.92 	&63.60 &	59.18 	&60.23 	&4.193
\\
\bottomrule
\end{tabular}}
\caption{Overall QA performance under different training strategies on CSL-daily-QA.}
\label{tab:knowledge_transfer_CSL_all}
\end{table*}

We further evaluate the proposed in-domain knowledge transfer strategy on
CSL-Daily-QA. The category-level and overall results are reported in
Tables~\ref{tab:knowledge_transfer_CSL}
and~\ref{tab:knowledge_transfer_CSL_all}, respectively. Compared with Stage 1
(CSLR pre-training), introducing Stage 2 (SLT pre-training) yields clear
improvements on the \emph{Non-Manual} question type, increasing F1 from
90.43 to 91.49 on the development set and from 92.15 to 93.19 on the test
set. Stage 2 also improves most translation metrics on the development set,
including BLEURT$_{\mathrm{SLT}}$, BLEU-2, BLEU-3, and BLEU-4, indicating
that SLT pre-training provides additional sentence-level semantic knowledge
and benefits language generation.

As shown in Table~\ref{tab:knowledge_transfer_CSL_all}, Stage 2 further
improves the overall QA performance on most lexical metrics. These improvements demonstrate that SLT pre-training
complements the sign-specific visual and temporal representations learned
during CSLR pre-training by strengthening cross-modal semantic alignment and
answer-generation capability.
 Nevertheless, the gains are not uniform across all categories. Gloss
recognition performance remains nearly unchanged, while structured QA
accuracy decreases slightly after Stage 2. Moreover, BLEURT and CIDEr show
small fluctuations on the test set. This suggests that CSLR pre-training
already provides a strong foundation for fine-grained recognition and
structural reasoning, whereas SLT pre-training mainly benefits
sentence-level semantics and generation-oriented metrics. Overall, the
results on CSL-Daily-QA confirm the complementary roles of the two training
stages: CSLR pre-training establishes sign-specific visual representations,
while SLT pre-training further enhances semantic alignment and language
generation.

\section{Qualitative Analysis}
\begin{CJK}{UTF8}{gbsn}

\begin{table*}[t]
\centering
\small
\begin{tabular}{p{4cm}|p{13cm}}
\toprule
\multicolumn{2}{c}{\textbf{PHOENIX14T-QA}}\\
\midrule
Question & Bestimmen Sie das finale Gloss-Token. \\
Ground Truth & Das Schluss-Gloss lautet ``REGEN''. \\
\hline
VideoLLaMA3-2B & Das Schluss-Gloss lautet ``GRAD''. \\

Qwen3-VL-2B-Instruct & Das Schluss-Gloss lautet ``REGEN''. \\

InternVL3-2B & Das Schluss-Gloss lautet ``REGEN''. \\

SLQAM (ours) & Das Schluss-Gloss lautet ``REGEN''. \\
\midrule
\multicolumn{2}{c}{\textbf{CSL-Daily-QA}}\\
\midrule
Question & 请确定序列末尾的终止手语词。 \\
Ground Truth & 序列结尾的gloss是“运动”。 \\
\hline
VideoLLaMA3-2B & 最后的gloss为“不”。 \\

Qwen3-VL-2B-Instruct & 最后的gloss为“可以”。 \\

InternVL3-2B & 序列结尾的gloss是“跑步”。 \\

SLQAM (ours) & 序列结尾的gloss是“运动”。 \\
\bottomrule
\end{tabular}
\caption{Qualitative comparison of \textbf{Position Reasoning} results on PHOENIX14T-QA and CSL-Daily-QA.}
\label{tab:SLQAQualitative1}
\end{table*}

\begin{table*}[t]
\centering
\small
\begin{tabular}{p{4cm}|p{13cm}}
\toprule
\multicolumn{2}{c}{\textbf{PHOENIX14T}}\\
\midrule
Question & Welches Gloss erscheint unmittelbar vor ``WIND''? \\
Ground Truth & ``WIND'' ist das erste Gloss; davor steht nichts. \\
\hline
VideoLLaMA3-2B & Vor ``WIND'' steht nichts mehr, es ist das erste Gloss. \\

Qwen3-VL-2B-Instruct & Vor ``WIND'' kommt ``SCHWACH''. \\

InternVL3-2B & Vor ``WIND'' kommt ``MAESSIG''. \\

SLQAM (ours) & ``WIND'' ist das erste Gloss; davor steht nichts. \\
\midrule
\multicolumn{2}{c}{\textbf{CSL-Daily-QA}}\\
\midrule
Question & 哪个gloss出现在“好处”之前？ \\
Ground Truth & “好处”的前一个gloss是“好”。 \\
\hline
VideoLLaMA3-2B & “好处”前面紧挨着“这”。 \\

Qwen3-VL-2B-Instruct & 在“好处”之前的是“有”。 \\

InternVL3-2B & “好处”前面紧挨着“有”。 \\

SLQAM (ours) & “好处”的前一个gloss是“好”。 \\
\bottomrule
\end{tabular}
\caption{Qualitative comparison of \textbf{Structural Reasoning} results on two benchmarks.}
\label{tab:SLQAQualitative2}
\end{table*}

\begin{table*}[t]
\centering
\small
\begin{tabular}{p{4cm}|p{13cm}}
\toprule
\multicolumn{2}{c}{\textbf{PHOENIX14T}}\\
\midrule
Question & Wann passiert ``SONNTAG'' zum letzten Mal? \\
Ground Truth & Das letzte Auftreten von ``SONNTAG'' ist an Position 1. \\
\hline
VideoLLaMA3-2B & Das letzte Auftreten von ``SONNTAG'' ist an Position 1. \\

Qwen3-VL-2B-Instruct & Das letzte Auftreten von ``SONNTAG'' ist an Position 2. \\

InternVL3-2B & Die letzte Stelle für ``SONNTAG'' ist Position 1. \\

SLQAM (ours) & Das letzte Auftreten von ``SONNTAG'' ist an Position 1. \\
\midrule
\multicolumn{2}{c}{\textbf{CSL-Daily-QA}}\\
\midrule
Question & “旁边”最后一次出现是在第几步？ \\
Ground Truth & “旁边”最近一次出现的位置是第2位。 \\
\hline
VideoLLaMA3-2B & “旁边”最后一次出现在第3位。 \\

Qwen3-VL-2B-Instruct & “旁边”最近一次出现的位置是第4位。 \\

InternVL3-2B & “旁边”最后一次出现于位置3。 \\

SLQAM (ours) & “旁边”最近一次出现的位置是第2位。 \\
\bottomrule
\end{tabular}
\caption{Qualitative comparison of \textbf{Visual Search} results on two benchmarks.}
\label{tab:SLQAQualitative3}
\end{table*}

\begin{table*}[t]
\centering
\small
\begin{tabular}{p{4cm}|p{13cm}}
\toprule
\multicolumn{2}{c}{\textbf{PHOENIX14T}}\\
\midrule
Question & Schreiben Sie die Gebärdenwörter, die Sie sehen, der Reihe nach auf. \\
Ground Truth & Die Gloss-Sequenz des Videos ist ``MORGEN WETTER WIE-AUSSEHEN DIENSTAG FUENFTE JANUAR''. \\
\hline
VideoLLaMA3-2B & Die Gloss-Sequenz des Videos ist ``MORGEN WETTER WIE-AUSSEHEN DIENSTAG DREIZEHN FEBRUAR''. \\

Qwen3-VL-2B-Instruct & Die Gloss-Sequenz des Videos ist ``MORGEN WETTER WIE-AUSSEHEN DIENSTAG VIERZEHN MAI''. \\

InternVL3-2B & Die Gloss-Sequenz des Videos ist ``MORGEN WETTER WIE-AUSSEHEN DIENSTAG NEUNTE FEBRUAR''. \\

SLQAM (ours) & Die Gloss-Sequenz des Videos ist ``MORGEN WETTER WIE-AUSSEHEN DIENSTAG FUENFTE JANUAR''. \\
\midrule
\multicolumn{2}{c}{\textbf{CSL-Daily-QA}}\\
\midrule
Question & 把看到的动作代号按顺序写下来。 \\
Ground Truth & 这个手语视频的gloss序列是“我 丈夫 在 美国 中国 菜 喜欢”。 \\
\hline
VideoLLaMA3-2B & 这个手语视频的gloss序列是“我 为 明天 跳舞 准备 音乐”。 \\

Qwen3-VL-2B-Instruct & 这个手语视频的gloss序列是“我 买 好了 面包 牛奶”。 \\

InternVL3-2B & 这个手语视频的gloss序列是“我 丈夫 每天 火锅 喜欢 吃”。 \\

SLQAM (ours) & 这个手语视频的gloss序列是“我 丈夫 在 美国 中国 菜 喜欢”。 \\
\bottomrule
\end{tabular}
\caption{Qualitative comparison of \textbf{Gloss Recognition} results on two benchmarks.}
\label{tab:SLQAQualitative4}
\end{table*}

\begin{table*}[t]
\centering
\small
\begin{tabular}{p{4cm}|p{13cm}}
\toprule
\multicolumn{2}{c}{\textbf{PHOENIX14T}}\\
\midrule
Question & Was ist der vollständige Satz, der vom Gebärdenden ausgedrückt wird? \\
Ground Truth & Diese Gebärdsequenz bedeutet ``am donnerstag regnet es vor allem im süden und osten mancherorts''. \\
\hline
VideoLLaMA3-2B & Diese Gebärdsequenz bedeutet ``am tag regnet es im süden und osten teilweise kräftig sonst ist es meist freundlich''. \\

Qwen3-VL-2B-Instruct & Diese Gebärdsequenz bedeutet ``am samstag im norden und westen viele wolken hier und da regnet oder schauert es''. \\

InternVL3-2B & Diese Gebärdsequenz bedeutet ``am sonntag ist es im süden freundlich''. \\

SLQAM (ours) & Diese Gebärdsequenz bedeutet ``am donnerstag regnet es vor allem im süden und südosten teilweise kräftig''. \\
\midrule
\multicolumn{2}{c}{\textbf{CSL-Daily-QA}}\\
\midrule
Question & 手语者表达的完整句子是什么？ \\
Ground Truth & 这段手语的意思是“你是我的客户,我应该请你吃饭.”。 \\
\hline
VideoLLaMA3-2B & 这段手语的意思是“他把手机里的自拍照都删掉了.”。 \\

Qwen3-VL-2B-Instruct & 这段手语的意思是“他把多余的钱都存到了银行.”。 \\

InternVL3-2B & 这段手语的意思是“你让我收拾房间,我怎么有种吃不消的感觉啊!”。 \\

SLQAM (ours) & 这段手语的意思是“你是我的客户,我应该请你吃饭.”。 \\
\bottomrule
\end{tabular}
\caption{Qualitative comparison of \textbf{Translation Understanding} results on two benchmarks.}
\label{tab:SLQAQualitative5}
\end{table*}

\end{CJK}

We present representative qualitative examples from the two constructed
\textsc{SignQA} datasets in Tables~\ref{tab:SLQAQualitative1}--\ref{tab:SLQAQualitative5},
covering \emph{Position Reasoning}, \emph{Structural Reasoning},
\emph{Visual Search}, \emph{Gloss Recognition}, and
\emph{Translation Understanding}. These examples provide a qualitative
comparison between general-purpose video-language models and our
sign-language-specific baseline.

For the reasoning-oriented tasks, general-purpose VLMs often produce
plausible but incorrect answers despite the relatively simple question structures. For example, in the \emph{Structural Reasoning} example, Qwen3-VL and
InternVL3 incorrectly predict a gloss before ``WIND'', although it is the
first gloss in the sequence. Similarly, in the \emph{Visual Search} example,
Qwen3-VL predicts an incorrect occurrence position for ``SONNTAG''. These
errors suggest that general-purpose VLMs may recognize relevant visual or
semantic content but struggle to accurately establish the fine-grained
temporal and sequential relationships required by SLQA. In contrast, our
model correctly identifies the queried positions and local relationships.

The \emph{Gloss Recognition} example further highlights the importance of
fine-grained sign-level visual understanding. While the general-purpose VLMs
produce gloss sequences with similar high-level content, they make errors in fine-grained details such as dates and numbers. These seemingly small
recognition errors can substantially affect the correctness of the entire
gloss sequence. Our model, benefiting from in-domain CSLR knowledge, correctly
recovers the complete gloss sequence in this example.

The difference is also evident in \emph{Translation Understanding}. The
general-purpose VLMs generate fluent German sentences, but their answers describe weather events that differ from those conveyed by the sign video.
This indicates that linguistic fluency alone does not guarantee accurate
sign-to-language semantic understanding. In contrast, our model preserves
the key semantic content of the video, including the temporal and spatial
information as well as the weather condition. Overall, these examples
illustrate that the proposed SLQA model is better able to connect fine-grained
visual evidence with sequential reasoning and language generation, rather
than merely producing plausible responses based on general video-language
priors.

\section{Impact Statement}

This work introduces a new task, Sign Language Question Answering (SLQA), together with two benchmark datasets and a baseline model. We believe this work has the potential to advance sign language understanding by enabling more flexible and interactive access to sign language videos. Compared with conventional sign language recognition and translation systems that produce fixed outputs, SLQA allows users to query different aspects of sign language videos through natural language, which may benefit applications such as sign language education, intelligent retrieval, and assistive communication technologies.
As with other data-driven sign language systems, our approach inherits the limitations and potential biases of the underlying datasets, including demographic imbalance, limited signer diversity, and constrained recording conditions. Consequently, the model may exhibit degraded performance for unseen signing styles, rare linguistic phenomena, or challenging visual environments. Therefore, the proposed system should not be deployed without appropriate human oversight in safety-critical or high-stakes applications where recognition errors could have significant consequences.
Overall, we believe that improving the flexibility and accessibility of sign language understanding can provide meaningful societal benefits, provided that the technology is developed and deployed responsibly.
 
\section{Limitations and Discussion}
We outline several limitations of the proposed method and discuss potential directions for future work.

\paragraph{Limited Dataset Coverage.}
The proposed SignQA datasets are constructed from PHOENIX14T and CSL-Daily, both of which focus on relatively constrained domains and controlled recording conditions. Therefore, they cannot fully capture the linguistic diversity, signer variability, and environmental complexity of sign language videos in real-world scenarios. As a result, models trained on these datasets may exhibit limited generalization to unseen domains, signing styles, and recording conditions. In future work, we plan to expand SignQA by incorporating larger and more diverse sign language datasets, with the goal of improving model robustness and enabling more reliable interaction with sign language videos in real-world applications.

\paragraph{Three-Stage Training.}
Our baseline model relies on a three-stage training pipeline to transfer in-domain knowledge from CSLR and SLT to SLQA. Although this strategy substantially improves performance, it inevitably increases the overall training time, computational cost, and implementation complexity. Developing a more efficient unified training strategy that can jointly learn sign recognition, translation, and question-answering capabilities remains an important direction for future research.

\end{document}